\documentclass{article}

\usepackage{microtype}
\usepackage{graphicx}
\usepackage{subcaption}
\usepackage{booktabs} 
\usepackage[table]{xcolor}
\definecolor{lightgray}{gray}{0.9}
\definecolor{lightred}{RGB}{255,230,230}

\usepackage{tabularx}

\usepackage{hyperref}

\usepackage[accepted]{icml2026}

\usepackage{amsmath}
\usepackage{amssymb}
\usepackage{mathtools}
\usepackage{amsthm}

\usepackage[capitalize,noabbrev]{cleveref}

\theoremstyle{plain}
\newtheorem{theorem}{Theorem}[section]

\newtheorem{corollary}[theorem]{Corollary}
\theoremstyle{definition}

\theoremstyle{remark}

\usepackage[textsize=tiny]{todonotes}

\icmltitlerunning{Scaling Unsupervised Multi-Source Federated Domain Adaptation through Group-Wise Discrepancy Minimization}

\begin{document}

\twocolumn[
  \icmltitle{Scaling Unsupervised Multi-Source Federated Domain Adaptation through Group-Wise Discrepancy Minimization}



  \icmlsetsymbol{equal}{*}

  \begin{icmlauthorlist}
    \icmlauthor{Larissa Reichart}{equal,TU}
    \icmlauthor{Cem Ata Baykara}{equal,TU}
    \icmlauthor{Ali Burak Ünal}{TU}
    \icmlauthor{Harlin Lee}{UNC}
    \icmlauthor{Mete Akgün}{TU}

  \end{icmlauthorlist}
    
  \icmlaffiliation{TU}{Department of Computer Science, Medical Data Privacy and Privacy-Preserving Machine Learning (MDPPML) Group and Institute for Bioinformatics and Medical Informatics (IBMI), University of Tübingen, Tübingen, Germany}
  \icmlaffiliation{UNC}{Department of Computer Science and the School of Data Science and Society, University of North Carolina at Chapel Hill, NC, USA}

  \icmlcorrespondingauthor{Cem Ata Baykara}{cem.baykara@uni-tuebingen.de}
  \icmlcorrespondingauthor{Larissa Reichart}{larissa.reichart@uni-tuebingen.de}

  \icmlkeywords{Machine Learning, ICML}

  \vskip 0.3in
]



\printAffiliationsAndNotice{\icmlEqualContribution}  

\begin{abstract}
    Unsupervised multi-source domain adaptation (UMDA) leverages labeled data from multiple source domains to generalize to an unlabeled target. While federated UMDA addresses privacy by avoiding raw data sharing, existing methods scale poorly as the number of sources increases, often suffering from high computational overhead or training instability. We propose GALA, a scalable and robust federated UMDA framework designed for high-diversity settings. GALA achieves scalability by coupling a novel inter-group discrepancy minimization objective that approximates pairwise alignment with linear complexity alongside a temperature-controlled, centroid-based weighting strategy for dynamic source prioritization. These components enable stable, parallelizable training across many heterogeneous sources, addressing a critical scalability bottleneck that remains largely unaddressed in current literature. To evaluate performance in high-diversity scenarios, we introduce Digit-18, a new benchmark comprising 18 datasets with varied synthetic and real-world domain shifts. Extensive experiments demonstrate that GALA achieves state-of-the-art results on standard benchmarks and significantly outperforms prior methods in large-scale settings where others either fail to converge or become computationally infeasible.
\end{abstract}

\section{Introduction} \label{introduction}

Unsupervised multi-source domain adaptation (UMDA) \citep{zhang2015multi} aims to learn a model that generalizes to an unlabeled target domain by leveraging labeled data from multiple sources. Unlike single-source adaptation, multi-source setups better reflect real-world conditions where data is naturally distributed across diverse environments. However, the presence of domain shifts among sources, in addition to the shift to the target, makes multi-source adaptation substantially more challenging.

Prior work has shown that alignment of source and target structures can improve robustness to distributional shift \citep{chang2019domain, zhao2020multi, dai2020adversarial, ganin2015unsupervised}. Yet in privacy-sensitive domains such as healthcare and finance, regulations like GDPR\footnote{General Data Protection Regulation, European Union} and CCPA\footnote{California Consumer Privacy Act} restrict data sharing and require both computation and data remain local. This makes centralized training infeasible and motivates the use of distributed UMDA approaches, such as federated \citep{konevcny2015federated, smith2017federated} or decentralized \citep{mcmahan2017communication} learning.

Current distributed UMDA methods are limited in scalability against diverse multi-source settings. Most are designed for small-scale scenarios involving only a handful of sources (typically 2–6)~\citep{kd3a,FACT,shot,fada}. As the number of source domains increases, these methods either require prohibitive computational costs or suffer from degraded performance and unstable convergence.

In this paper, we propose \textbf{G}rouping-based \textbf{A}daptive \textbf{L}e\textbf{a}rning (\textbf{GALA}), a federated UMDA framework designed to scale with the heterogeneity of diverse source domains. GALA combines two key components: (1) an inter-group discrepancy minimization objective that aligns aggregated source predictions without computing all pairwise discrepancies; and (2) a temperature-scaled centroid-based weighting scheme that dynamically estimates each source’s alignment with the target domain. By randomly partitioning sources into groups and minimizing disagreement between their weighted predictions on the unlabeled target domain, GALA approximates the global alignment objective in a scalable and robust manner.

To evaluate high-source settings more realistically than duplicating domains across clients, we introduce Digit-18, a new benchmark of 18 digit datasets spanning diverse synthetic and real-world shifts. Extensive experiments show that GALA not only matches or exceeds state-of-the-art performance on standard UMDA benchmarks, but also maintains stability and robustness as the number of diverse sources grows, where existing methods fail to converge or become computationally infeasible. Our Contributions are as follows:

\begin{itemize}
    \item To our knowledge, we are the first to explicitly address scalability in distributed UMDA with respect to the number of heterogeneous source domains, showing that existing approaches either fail to converge or become computationally infeasible as the number of sources grows.
    
    \item We propose Inter-Group Discrepancy, a group-level discrepancy objective that approximates the full pairwise disagreement at linear cost with low variance, enabling robust alignment when source count is large.
    
    \item We introduce GALA, a scalable federated UMDA algorithm that couples IGD with a temperature-scaled, centroid-based similarity weighting to prioritize target-relevant sources and mitigate negative transfer in high-diversity settings.
    
    \item We release \textbf{Digit-18}, a challenging large-scale UMDA benchmark designed to test scalability and source heterogeneity, and use it alongside standard datasets for comprehensive evaluation.
    
    \item Through extensive experiments and ablation studies, we demonstrate that GALA consistently improves accuracy, stability, and convergence compared to prior methods across both standard and high-diversity multi-source scenarios.
\end{itemize}

\section{Related Work} \label{relatedWork}

\paragraph{Unsupervised Multi-Source Domain Adaptation} 
Standard UMDA techniques aim to learn domain-invariant representations that generalize well to an unlabeled target domain by reducing discrepancies between the source and target distributions \citep{ben2010theory,NEURIPS2018_717d8b3d}. This is mainly achieved by two approaches: maximum mean discrepancy (MMD) \citep{tzeng2014deep, Digit5} and adversarial training \citep{saito2018maximum, fada, liu2018unified}.

\paragraph{Federated Domain Adaptation}
Federated learning is a distributed machine learning technique which allows collaborative training of a global model through aggregation of local model updates \citep{konevcny2016federated}. Federated UMDA was first proposed by \citet{fada}, which uses adversarial training to minimize $\mathcal{H}$-divergence without direct access to data. 
FACT \citep{FACT} is a recent approach that aligns source and target representations using inter-domain differences rather than adversarial training. It achieves state-of-the-art performance on digit datasets while being inherently scalable and efficient.
However, while scalable, this approach introduces high variance and suffers from convergence issues when the number of sources grows, as each training step involves only a single pair of sources. Our empirical results show that FACT’s performance becomes unstable in high-source scenarios and often fails to converge on challenging target domains.

\paragraph{Decentralized Domain Adaptation} 

Decentralized UMDA methods resemble their federated counterparts, with the key difference that training is not coordinated through a central server \citep{wu2021collaborative}. \citet{shot} proposes a source-free strategy for single-source domain adaptation that can also be extended to multi-source settings. In SFDA \citep{SFDA}, the Multi-Domain Model Generalization Balance (MDMGB) algorithm is introduced to adaptively weight multiple source models according to their similarity to the target domain. The target predictor is then trained separately using pseudo-labeling and information maximization. However, while it is efficient in terms of communication rounds, its performance on benchmark datasets is not state-of-the-art. 

More recently, \citet{fdfm} propose to fuse domain-invariant and domain-specific features, arguing that retaining domain-specific information is important for effective classification. Currently, \citet{kd3a} represents the state-of-the-art in decentralized UMDA, using a consensus-driven alignment strategy that achieves strong accuracy and robustness against negative transfer across multiple benchmarks.

However, their method, KD3A, is inherently not scalable to high-source settings, as it requires per-domain optimization and divergence computation to be performed locally at the target. This makes it unsuitable for distributed scenarios where target access is limited or expensive. Our empirical analysis shows that even when source training is parallelized to mimic practical deployment, KD3A’s computation time grows exponentially with the number of sources, becoming infeasible in large multi-source settings.

\section{Methodology} \label{methodology}

\textbf{Preliminaries.} \quad  
In a UMDA setting, we have $N$ distinct source domains $\{\mathbb{D}_{S}^n\}_{n=1}^N$ where each domain contains $K_n$ labeled samples as $\{\mathbb{D}_{S}^n\} :=\{(x_i^n, y_i^n)\}_{i=1}^{K_n}$, and a target domain $\mathbb{D}_T$ with $K_T$ unlabeled samples $\mathbb{D}_{T} :=\{x_i^T\}_{i=1}^{K_T}$. We consider a $C$‐way classification task shared across all domains, and assume every domain contains samples from every class. The main objective of UMDA is to learn a \emph{feature extractor} $G:\mathcal X\to\mathbb R^d$, and a \emph{predictor} $F: \mathbb{R}^d \to \Delta^C$, where $\Delta^C$ is the probability vector over $C$ classes. Together they define the model $h = F\circ G \in \mathcal{H}$ that minimizes the task error 
$\epsilon_{\mathbb{D}_T}(h) \;=\; \Pr_{(x,y)\sim\mathbb{D}_T}\bigl[h(x)\neq y\bigr]$. See \cite{ben2010theory,NEURIPS2018_717d8b3d} for formal definitions of $\mathcal{H}$-divergences $d_\mathcal{H}, d_{\mathcal H\Delta\mathcal H}$.

\subsection{Federated UMDA Theoretical Motivation} \label{sec:umda}


For any set of weights $\{w_n\}$ that determines the contribution of each source domain to the final predictor, the following generalization bound holds for federated UMDA. It stems from a simple convex combination of the classical UMDA generalization bound  \cite{NEURIPS2018_717d8b3d,ben2010theory}, so we do not make claims on novelty or tightness. We present this only to gain some insight into federated UMDA with large data heterogenity and motivate our approach.
\begin{corollary}[Generalization bound of federated UMDA]\label{thm:multisource}
Let $w_1,\dots,w_N\in\mathbb{R}_+$ satisfy $\sum_{n=1}^N w_n=1$. $m$ is sample size of $T$, $\tilde{S}$ is pooled $S$, and $p_n$ is relative sample size of $S_n$ compared to $\tilde{S}$. Then for any $h\in \mathcal{H}$, w.h.p.
\begin{align}
  &  \epsilon_{D_T}(h)
  \;\le\;\hat\epsilon_{\tilde S}(h)+\sum_n(w_n- p_n)\hat\epsilon_{D_S^n}(h) \\
  +&\boxed{\sum_{n=1}w_n\;\tfrac12\,d_{ H\Delta H}\bigl(D_S^n,D_T\bigr)} +\sum_{n=1}w_n\lambda_n + \tilde{O}\left(\frac 1 m\right),\nonumber
  \end{align}
where $\lambda_n:=\min_h  \epsilon_{D_S^n}(h)+\epsilon_{D_T}(h)$.
\end{corollary}
This highlights the effect of federated learning. The first term is the empirical loss one would obtain if all source data were pooled, and the second term isolates the deviation introduced by source heterogeneity and the use of $w_n$ instead of sample-proportion weights, becoming positive or negative depending on whether clients contribute unevenly relative to their dataset sizes. 

This also makes explicit that the optimal joint error term depends on $w_n$, so the bound cannot be directly minimized in practice to identify the ``best'' $w_n$ or single source. However, the boxed divergence term does suggest that aligning source feature distributions with the target may lead to reduced target error and can be attempted without labels. GALA fits in this line of work. In the following sections, we discuss designing and computing a scalable proxy loss $\mathcal{L}_{IGD}$ for the boxed divergence term.

\subsection{Prior Work: Minimizing Divergence for UMDA}

Adversarial domain adaptation typically achieves alignment between source and target feature distributions by training a domain discriminator to make source and target feature distributions indistinguishable \citep{NEURIPS2018_717d8b3d}. However, \citet{zhao2019learning, liu2019transferable} warn that aligning only input features can be insufficient or even harmful and feature-level discrepancy minimization does not necessarily yield invariant representations and may increase the target error in some cases. In federated multi-source settings this can be more problematic because discriminator training generally requires simultaneous access to samples from multiple domains. 

Instead, \citet{FACT} recently proposed a non-adversarial and federated alternative that measures predictor disagreement on unlabeled target samples to quantify inter-source differences and minimizes the average pairwise disagreement between source predictors. This approach allows identifying domain-specific artifacts without explicit adversarial maximization. This idea is captured by the full pairwise discrepancy:

\begin{equation}\label{eq:l_full}
    \mathcal{L}_{\mathrm{full}} = \sum_{i<j}\mathbb{E}_{x \sim \mathbb{D}_T} \left[\|F_i(G(x)) - F_j(G(x))\|_1\right],
\end{equation}

where the sum $\sum_{i < j}$ ranges over all unordered pairs of distinct source domains $i, j \in \{1, \dots, N\}$. Because this formulation scales quadratically with the number of sources, requiring $\binom{N}{2}$ pairwise comparisons, it becomes impractical in large-scale settings. To address this, \citet{FACT} further introduces the concept of Inter-Domain Distance (IDD) minimization, where, at each round, two source domains are randomly selected, and the disagreement between their predictors on target data is minimized. \citet{FACT} formally define this as:

\begin{equation} \label{eq:idd}
    \mathcal{L}_{\text{IDD}}^{(i,j)} = \mathbb{E}_{x\sim \mathbb{D}_T}\left[\left\|F_i(G(x)) - F_j(G(x))\right\|_1\right],
\end{equation}

which encourages domain-invariant representations by aligning the outputs of individual source predictors on the target distribution. Although efficient and unbiased, this approach introduces high variance, as each update reflects only the behavior of a single random pair of sources. Our empirical findings show that this approach becomes unstable in diverse multi-source settings and frequently fails to converge on challenging target domains.

An intuitive explanation of the connection between Corollary \ref{thm:multisource} and the proxy losses are deferred to Appendix \ref{app:connection}.

\subsection{Achieving Scalability through Inter-Group Discrepancy} \label{sec:inter-group-discrpency}

To mitigate these issues, we introduce Inter-Group Discrepancy (IGD), a group-level discrepancy objective that serves as an efficient, low-variance approximation to $\mathcal{L}_{\mathrm{full}}$ without requiring full pairwise computation or reliance on a single random pairing. At each training round, IGD randomly partitions the $N$ source predictors $\{F_n\}_{n=1}^N$ into two disjoint groups $\mathcal{G}_1$ and $\mathcal{G}_2$, forms weighted average predictors for each group, and minimizes their $\ell_1$ disagreement on unlabeled target data:

\begin{equation} \label{eq:igd}
    \mathcal{L}_{\mathrm{IGD}} = \mathbb{E}_{x \sim \mathbb{D}_T}\left[ \| F_{\mathcal{G}_1}(G(x)) - F_{\mathcal{G}_2}(G(x)) \|_1 \right],
\end{equation}

with group predictors:

\begin{equation}\label{eq:group_predictors}
    F_{\mathcal{G}_1} = \sum_{n \in \mathcal{G}_1} \tilde{w}_n F_n, \quad F_{\mathcal{G}_2} = \sum_{n \in \mathcal{G}_2} \tilde{w}_n F_n.
\end{equation}


\paragraph{Theoretical Intuition.}
IGD and IDD both aim to estimate $\mathcal{L}_{\mathrm{full}}$, which is expensive to compute. IDD is an efficient and unbiased but high-variance estimator of $\mathcal{L}_{\mathrm{full}}$. As for IGD, a quick algebraic manipulation of Eq. \eqref{eq:igd} and \eqref{eq:group_predictors} along with $\sum_{i\in G_k}\tilde w_i = 1$ and $\tilde w_i \ge 0$ shows that the IGD objective is equivalent to

\begin{equation*}
   \mathbb{E}_{x\sim\mathbb{D}_T}\Bigg[\Big\|\sum_{i\in G_1}\sum_{j\in G_2}\tilde w_i\tilde w_j\big(F_i(G(x))-F_j(G(x))\big)\Big\|_1\Bigg].
\end{equation*}

By averaging many pairwise differences before taking the $\ell_1$ norm, IGD yields lower variance than the single-pair discrepancy estimator such as the IDD term in Eq.~\ref{eq:idd}. Moreover, Jensen's inequality gives the useful bound

\begin{equation}
    \mathcal{L}_{\mathrm{IGD}} \le \sum_{i\in G_1}\sum_{j\in G_2}\tilde w_i\tilde w_j\, L_{\mathrm{IDD}}^{(i,j)} \le \max_{i,j} L_{\mathrm{IDD}}^{(i,j)}.
\end{equation}

Taking expectation over random balanced splits shows that, under uniform (or non-adversarial) weights, IGD is a biased approximation of the full pairwise objective (see Appendix \ref{app:intuition} for a short derivation). In practice, this bias is small, and the resulting variance reduction leads to more stable training and improved final performance in highly diverse multi-source settings, as supported by our extensive experiments.

\subsection{Enhancing IGD Through Source Relevance}

While IGD provides a low-variance approximation to the full pairwise objective under uniform weights, its practical effectiveness depends on how much influence each source exerts in the group aggregates. If group predictors place large importance on irrelevant or noisy sources, the aggregated outputs can be biased and drive negative transfer rather than alignment. Therefore, in addition to minimizing group disagreement, we require a lightweight and private mechanism that (i) estimates a global relevance score \(w_n\) for each source (so that well-aligned sources contribute more to the final predictor), and (ii) produces group-normalized weights \(\tilde w_n\) (Eq.~\ref{eq:group_predictors}) so that IGD focuses its alignment on relevant sources within each partition.

\paragraph{Sources Weighting via Centroid Similarity.} To estimate similarity between each source domain and the target without accessing labels or sharing data, we adopt a centroid-based proxy inspired by the MDMGB algorithm of \citet{SFDA}. MDMGB computes similarity using class-wise centroids in feature space. Specifically, each domain computes a soft centroid for class $c$ as:

\[
r_n^c = \frac{\sum_{x \in \mathbb{D}_S^n} \delta_c(x) \cdot G(x)}{\sum_{x \in \mathbb{D}_S^n} \delta_c(x)}, \quad
r_T^c = \frac{\sum_{x \in \mathbb{D}_T} \delta_c(x) \cdot G(x)}{\sum_{x \in \mathbb{D}_T} \delta_c(x)},
\]

where $\delta_c(x)$ is the softmax probability for class $c$ from predictor $F$. A cosine similarity score between source and target centroids is computed:

\begin{equation} \label{eq:domain_similarity}
    S(r_T^c, r_n^c) = \sum_{c=1}^C \frac{\langle r_T^c, r_n^c \rangle}{\|r_T^c\| \|r_n^c\|} + 1.
\end{equation}

\paragraph{Limitations of MDMGB in Diverse Source Settings.} While effective in low-diversity settings, the original MDMGB approach underperforms when source domains vary widely in quality or distribution. In such cases, the computed similarities fail to sharply penalize misaligned domains, resulting in negative transfer and poor target alignment. Our ablation analysis show that using unmodified MDMGB in diverse multi-source scenarios performs comparably to using uniform weights.

To address this limitation, we propose a modified version of MDMGB, which we refer to as MDMGB+. This variant introduces a softmax-based selection mechanism with a tunable temperature parameter $\tau > 0$ that sharpens the similarity contrast among source domains. We define the global relevance score for each source as:

\begin{equation} \label{eq:mdmgb+}
    w_n = \frac{\exp(\tau \cdot S(r_T^c, r_n^c))}{\sum_{j=1}^N \exp(\tau \cdot S(r_T^c, r_j^c))}.
\end{equation}

Within each partition \(\mathcal G_k\), these global scores are re-normalized to form group-specific weights:

\begin{equation} \label{eq:normalized_weights}
\tilde{w}_n =
\frac{\exp\!\big(\tau \cdot S(r_T^c, r_n^c)\big)}
{\sum_{i \in \mathcal{G}_k} \exp\!\big(\tau \cdot S(r_T^c, r_i^c)\big)},
\quad n \in \mathcal{G}_k,\; k \in \{1,2\}.
\end{equation}

The temperature parameter \(\tau\) controls the selectivity of the weighting: higher values amplify small similarity differences, assigning more importance to sources better aligned with the target. MDMGB+ enables IGD to remain effective in diverse multi-source settings, where relevant domains might otherwise be dominated by dissimilar ones.

\subsection{The GALA Algorithm}

\begin{figure*}[ht!]
    \centering
    \includegraphics[width=0.9\textwidth]{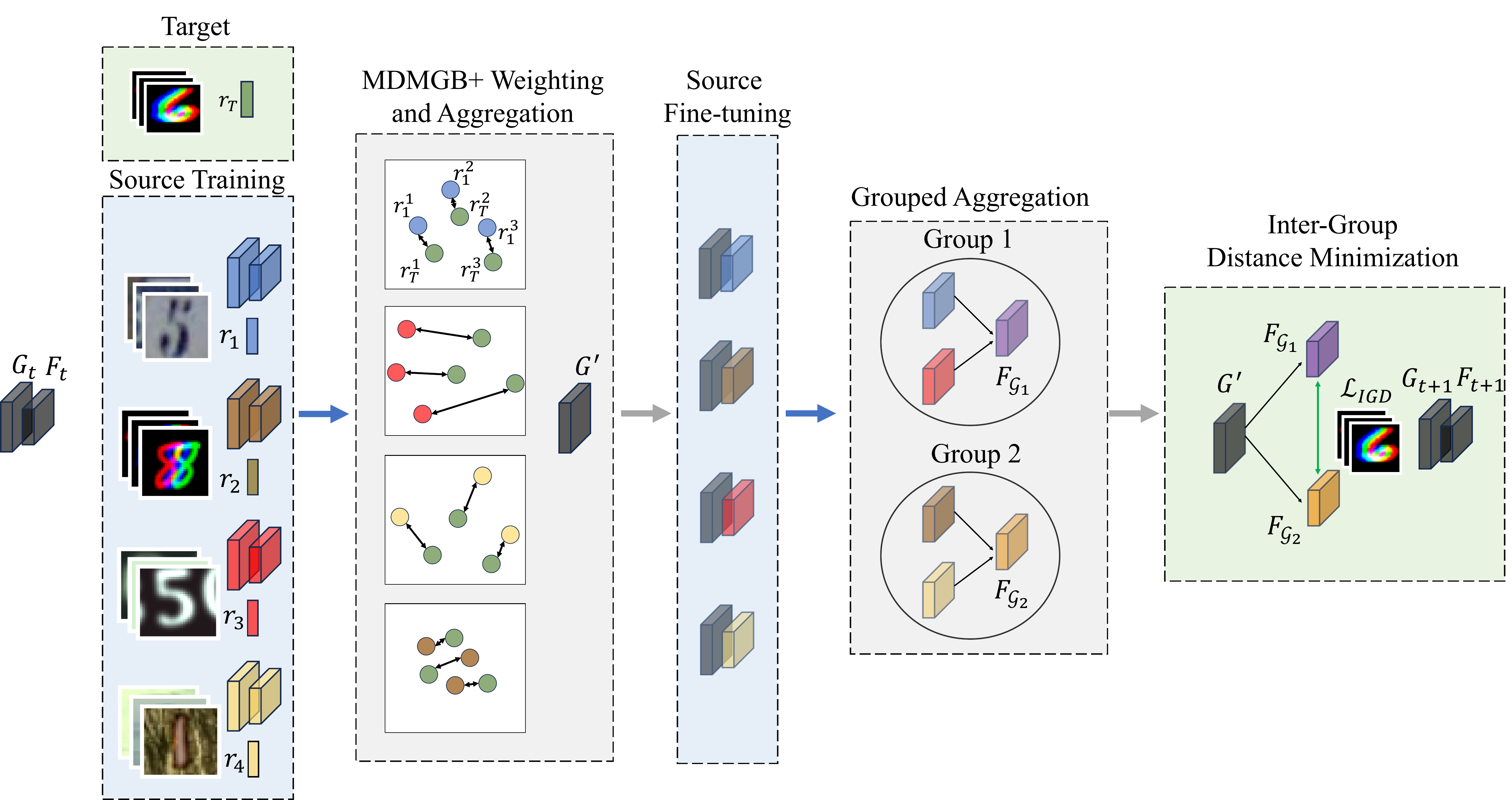}
    \caption{Visual overview of the GALA framework for federated unsupervised multi-domain adaptation.}
    \label{fig:method_visualization}
\end{figure*}

GALA combines the IGD objective and MDMGB+ relevance weighting into a federated process that iteratively aligns source predictors with an unlabeled target. A high-level overview is shown in Figure \ref{fig:method_visualization}. At round $t$, the current global model is $h_t = F_t\circ G_t$. Source and target clients compute class-wise soft centroids in feature space ($r_n^c$ and $r_T^c$) from local data, and share only these centroid summaries with the server. Clients also perform local training updates of the global model and transmit the updated parameters to the server.

Using the received centroids, the server computes a global relevance profile $\{w_n\}$ via the MDMGB+ softmax (Eq.~\ref{eq:mdmgb+}) which differentiates between well-aligned and misaligned sources and yields group-normalized weights $\{\tilde w_n\}$. The relevance scores are used to aggregate the client's feature extractors to a consensus feature extractor $G'$. The corresponding predictors are returned to the clients for local fine-tuning to maintain compatibility with $G'$. 

To align the consensus extractor with the unlabeled target, the fine-tuned predictors are randomly partitioned into two groups to form group predictors $F_{\mathcal G_1}, F_{\mathcal G_2}$ (Eq.~\ref{eq:group_predictors}). The extractor is then updated on the target data by minimizing the IGD loss (Eq.~\ref{eq:igd}) to obtain $G''$. Finally, the group predictors are merged into a single relevance-weighted global predictor $F_{t+1}$, and the next global model is defined as $h_{t+1} = F_{t+1}\circ G_{t+1}$ with $G_{t+1} = G''$. This cycle repeats until convergence. The step-by-step algorithm pseudo-code and implementation details are provided in the Appendix.

\section{Experiments} \label{experiments}

\begin{table*}[ht]
    \centering
    \caption{UMDA accuracy (\%) on the Digit-Five dataset.}
    \renewcommand{\arraystretch}{1.2}
    \begin{tabular}{l|ccccc|c}
        \hline
        \rowcolor{lightgray} Methods & \textit{mnist} & \textit{mnistm} & \textit{svhn} & \textit{syn} & \textit{usps} & Avg \\
        \hline
        Oracle & $99.5_{\pm0.08}$ & $95.4_{\pm0.15}$ & $92.3_{\pm0.14}$ & $98.7_{\pm0.04}$ & $99.2_{\pm0.09}$ & 97.0 \\
        Source-only & $92.3_{\pm0.91}$ & $63.7_{\pm0.83}$ & $71.5_{\pm0.75}$ & $83.4_{\pm0.79}$ & $90.71_{\pm0.54}$ & 80.3 \\
        \hline
        MDAN & $97.2_{\pm0.98}$ & $75.7_{\pm0.83}$ & $82.2_{\pm0.82}$ & $85.2_{\pm0.58}$ & $93.3_{\pm0.48}$ & 86.7 \\
        $M^3$SDA & $98.4_{\pm0.68}$ & $72.8_{\pm1.13}$ & $81.3_{\pm0.86}$ & $89.6_{\pm0.56}$ & $96.2_{\pm0.81}$ & 87.7 \\
        CMSS & $99.0_{\pm0.08}$ & $75.3_{\pm0.57}$ & $88.4_{\pm0.54}$ & $93.7_{\pm0.21}$ & $97.7_{\pm0.13}$ & 90.8 \\
        DSBN & 97.2 & 71.6 & 77.9 & 88.7 & 96.1 & 86.3 \\
        \hline
        FADA & $91.4_{\pm0.7}$ & $62.5_{\pm0.7}$ & $50.5_{\pm0.3}$ & $71.8_{\pm0.5}$ & $91.7_{\pm1}$ & 73.6 \\
        SHOT & $98.2_{\pm0.37}$ & $80.2_{\pm0.41}$ & $84.5_{\pm0.32}$ & $91.1_{\pm0.23}$ & $97.1_{\pm0.28}$ & 90.2 \\
        SFDA & $99.1$ & $72.3$ & $86.0$ & $90.4$ & $98.1$ & $89.2$ \\
        KD3A & $99.2_{\pm0.12}$ & $87.3_{\pm0.23}$ & $85.6_{\pm0.17}$ & $89.4_{\pm0.28}$ & $\mathbf{98.5_{\pm0.25}}$ & 92.0 \\
        FACT & $\mathbf{99.3_{\pm0.12}}$ & $91.4_{\pm0.53}$ & $90.9_{\pm0.40}$ & $94.8_{\pm0.22}$ & $98.3_{\pm0.11}$ & 95.0 \\
        \rowcolor{lightred} GALA & $99.2_{\pm0.05}$ & $\mathbf{93.0_{\pm0.43}}$ & $\mathbf{91.2_{\pm0.16}}$ & $\mathbf{95.2_{\pm0.17}}$ & $98.3_{\pm0.10}$ & \textbf{95.4}  \\
        \hline
        
    \end{tabular}
    
    \label{tab:digit-Five}
\end{table*}

\begin{table}[ht]
\centering
\caption{UMDA accuracy (\%) on the Office-Caltech10.}
\label{tab:office_caltech10}
\renewcommand{\arraystretch}{1.2}
\resizebox{\columnwidth}{!}{%
\begin{tabular}{l|cccc|c}
\hline
\rowcolor{lightgray} Methods & \textit{amazon} & \textit{caltech} & \textit{dslr} & \textit{webcam} & Avg \\
\hline
Oracle         & 99.7 & 98.4 & 99.8 & 99.7 & 99.4 \\
Source-only    & 86.1 & 87.8 & 98.3 & 99.0 & 92.8 \\
\hline
MDAN           & 98.9 & 98.6 & 91.8 & 95.4 & 96.1 \\
M$^3$SDA       & 94.5 & 92.2 & 99.2 & 99.5 & 96.4 \\
CMSS           & 96.0 & 93.7 & 99.3 & 99.6 & 97.2 \\
DSBN           & 93.2 & 91.6 & 98.9 & 99.3 & 95.8 \\
DANE           & 97.4 & 97.3 & 100.0 & 100.0 & 98.7 \\
\hline
FADA           & 84.2\scriptsize{$\pm$0.5} & 88.7\scriptsize{$\pm$0.5} & 87.1\scriptsize{$\pm$0.6} & 88.1\scriptsize{$\pm$0.4} & 87.1 \\
SHOT           & 96.4 & 96.2 & 98.5 & 99.7 & 97.7 \\
FACT           & 96.3 & 95.5 & 99.4 & 99.0 & 97.6 \\
KD3A           & \textbf{97.4\scriptsize{$\pm$0.08}} & \textbf{96.4\scriptsize{$\pm$0.11}} & 98.4\scriptsize{$\pm$0.08} & 99.7\scriptsize{$\pm$0.02} & \textbf{97.9} \\
\rowcolor{lightred} GALA   & 96.5\scriptsize{$\pm$0.19} & 95.0\scriptsize{$\pm$0.17} & \textbf{100.0\scriptsize{$\pm$0.00}} & \textbf{99.8\scriptsize{$\pm$0.17}} & 97.8 \\
\hline
\end{tabular}
}
\end{table}

We evaluate GALA on standard UMDA benchmarks, study its scalability as the number of distinct source domains increases, and analyze its training efficiency in federated settings. Our evaluation considers three datasets:

\paragraph{Digit-Five.} \citep{Digit5} A standard benchmark with five digit domains and moderate source-target shifts.

\paragraph{Office-Caltech10.} \citep{OfficeDataset, Caltech256} A small-scale object recognition benchmark with four domains and 10 shared categories.

\paragraph{Digit-18 (ours).} A new large-scale benchmark comprising 18 diverse digit domains. These domains were created by systematically applying techniques such as background augmentation, scaling, and color channel stacking to existing digit datasets, resulting in substantial distributional shifts. Full details on dataset generation and inter-domain similarity analysis are provided in the Appendix.

\textbf{Baselines.} We compare GALA against both centralized and federated UMDA baselines. Central baselines include MDAN \citep{NEURIPS2018_717d8b3d}, M$^3$SDA \citep{Digit5}, CMSS \citep{yang2020curriculum}, DSBN \citep{chang2019domain}, and DANE \citep{dane}. Distributed baselines include SHOT \citep{shot}, FADA \citep{fada}, SFDA \citep{SFDA}, FACT \citep{FACT}, and KD3A \citep{kd3a}. We also report a Source only baseline, which is the model trained only on labeled source domains and evaluated on the target without any adaptation, and an Oracle baseline, which is the model trained and validated with access to labeled target domain data.


\textbf{Implementation details.} For digit datasets (Digit-Five and Digit-18), we use a 2-layer CNN with two convolutional blocks followed by three fully connected layers with dropout and batch normalization. For Office-Caltech10, we adopt a ResNet101 pretrained on ImageNet, followed by a two-layer MLP predictor (see Appendix for full architecture). All models are trained with SGD (momentum 0.9, weight decay $5 \times 10^{-4}$). The temperature parameter $\tau$ is set to $1.0$ for Office-Caltech10 and Digit-18, and to $0.2$ for Digit-Five. For Digit-Five and Digit-18, we use a custom learning rate scheduler with decay factor $\gamma = 0.75$, and for Office-Caltech10, exponential decay with $\gamma = 0.9$. The batch size is set to 128, and all models are trained for 500 epochs. Communication occurs once per epoch ($r=1$), using a single epoch for each stage: source training, source fine-tuning, and inter-group discrepancy minimization. We report mean ± std accuracy over five runs, using an AMD EPYC 7713 CPU and NVIDIA A100 GPU (40GB).


\subsection{Performance on Standard Benchmarks}

\begin{table*}[ht]
\centering
\scriptsize
\caption{
Accuracy (\%) on various target domains using the Digit-18 benchmark.
}
\begin{tabularx}{\textwidth}{l|*{9}{>{\centering\arraybackslash}X}|>{\centering\arraybackslash}X}
\hline
\rowcolor{lightgray} Method & \textit{mnist} & \textit{mnistm} & \textit{svhn} & \textit{syn} & \textit{usps} & \textit{synm} & \textit{svhn-xs} & \textit{svhnstack} & \textit{usps-m} & Avg \\
\hline
Oracle & $99.0_{\pm0.03}$ & $95.6_{\pm0.26}$ & $88.4_{\pm0.15}$ & $97.0_{\pm0.12}$ & $98.9_{\pm0.12}$ & $83.8_{\pm0.32}$ & $84.7_{\pm0.37}$ & $86.5_{\pm0.04}$ & $92.0_{\pm0.51}$ & $91.8$ \\
\hline
FACT & $98.6_{\pm0.20}$ & $87.6_{\pm1.58}$ & $92.5_{\pm0.41}$ & $97.4_{\pm0.46}$ & $98.3_{\pm0.26}$ & $79.4_{\pm1.51}$ & $79.1_{\pm4.75}$ & $91.9_{\pm1.63}$ & $87.6_{\pm1.06}$ & $87.6$ \\
\rowcolor{lightred} GALA & $\mathbf{99.3}_{\pm0.07}$ & $\mathbf{95.2}_{\pm0.10}$ & $\mathbf{95.4}_{\pm0.17}$ & $\mathbf{98.4}_{\pm0.05}$ & $\mathbf{99.0}_{\pm0.10}$ & $\mathbf{85.8}_{\pm0.23}$ & $\mathbf{88.6}_{\pm0.29}$ & $\mathbf{94.8}_{\pm0.10}$ & $\mathbf{92.9}_{\pm0.34}$ & $\mathbf{92.9}$ \\
\hline
\end{tabularx}
\label{tab:digit18_performance}
\end{table*}

\paragraph{Digit-Five.} 
As shown in Table~\ref{tab:digit-Five}, GALA achieves the highest average accuracy on Digit-Five and performs best on all but two target domains. On the remaining targets, it is only marginally below the best method with differences of around 0.2\%. These results demonstrate that GALA matches or surpasses state-of-the-art methods on this standard benchmark, serving as a strong baseline for the more challenging settings considered next.

\paragraph{Office-Caltech10.} On this small benchmark, GALA ranks second among distributed methods with 97.8\% accuracy, closely matching KD3A (97.9\%). While KD3A is slightly better on Amazon and Caltech, GALA achieves the highest accuracy on DSLR and Webcam, including 100\% on DSLR.

\subsection{Scalability Under Growing Source Diversity}

Next, we evaluate performance as the number of source domains increases. Starting from the 4-source Digit-Five setup, we progressively add Digit-18 domains by increasing task difficulty (lowest self-performance first; see Appendix). Figure~\ref{fig:source_client_plot} shows results across Digit-Five targets.

\begin{figure}[ht]
    \centering
    \includegraphics[width=\linewidth]{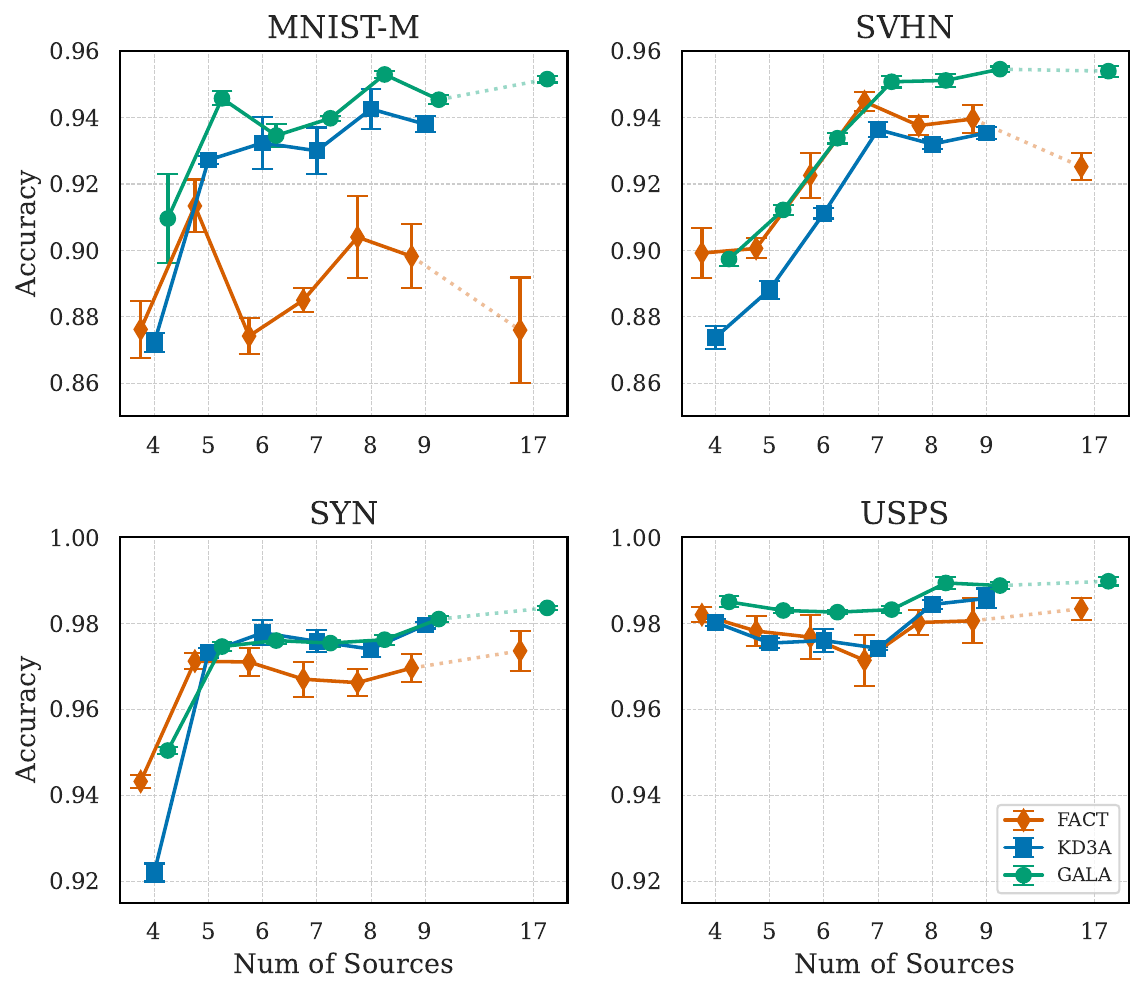}
    \caption{Performance across Digit-Five targets for increasing source domains. KD3A is excluded beyond 9 sources due to exponential runtime.}
    \label{fig:source_client_plot}
\end{figure}

While all methods initially benefit from additional source domains, performance diverges as dissimilar or noisy sources are added. FACT becomes unstable beyond 9 sources, showing higher variance due to its reliance on randomly sampled pairs. KD3A remains robust but suffers exponential runtime growth, making it impractical for high-source settings. In contrast, GALA maintains stable accuracy with dynamic weighting that suppresses negative transfer. Across all targets, GALA improves with more sources and consistently outperforms KD3A and FACT. Due to KD3A’s runtime, the full 18-domain evaluation is restricted to FACT and GALA.

\begin{figure*}[ht]
    \centering
    \includegraphics[width=\linewidth]{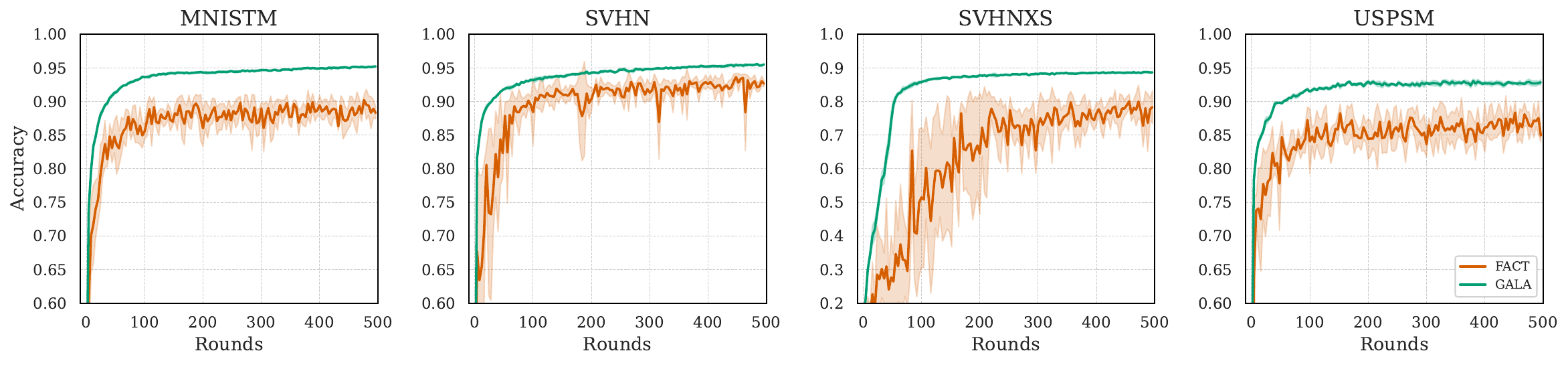}
    \caption{ Test accuracy over training rounds for four Digit-Five target domains in the full Digit-18 setup.}

    \label{fig:digit18_performance_1}
\end{figure*}



\paragraph{Full Digit-18 Results.} 

Table~\ref{tab:digit18_performance} reports accuracy across 9 target domains in the full 18-source setting, highlighting the challenges of this high-diversity scenario. GALA achieves a 5.3\% average gain over FACT and outperforms all baselines on every target. Improvements are particularly pronounced on the most challenging domains, such as SVHNXS (+9.5\%) and SYNM (+6.4\%), where FACT struggles but GALA still maintains strong performance. These results show that in multi-source settings with more than ten sources, GALA overcomes prior limitations and establishes a new state-of-the-art across all targets.

\paragraph{Training Dynamics.} 
Figure~\ref{fig:digit18_performance_1} shows test accuracy over training rounds for four challenging target domains. GALA converges quickly, surpassing FACT’s final accuracy within the first 100 rounds, and continues to improve slightly thereafter. Accuracy remains stable both between rounds and across 10 runs. In contrast, FACT exhibits large fluctuations and struggles on difficult targets such as MNISTM, SVHN, and SVHNXS. Additional results on easier targets are provided in the appendix, where the same trend appears but is less pronounced.

\subsection{Runtime Comparison} \label{runtime_comparison}

To evaluate computational efficiency, we compare per-round runtimes under an idealized federated setting where client-side operations are parallelized to simulate practical execution without bandwidth constraints (Table~\ref{tab:runtime}).

\begin{table}[ht]
\centering
\caption{Per-round training time (in seconds) for varying numbers of source domains.}
\renewcommand{\arraystretch}{1.2}
\resizebox{\columnwidth}{!}{%
\begin{tabular}{c|cccc}
\hline
\rowcolor{lightgray} \# Sources & 3 & 5 & 7 & 9 \\
\hline
KD3A   & $50.73_{\pm 0.3}$ & $216.03_{\pm 2.7}$ & $1029.84_{\pm 10.6}$ & $5600.48_{\pm 32.0}$ \\
FACT   & $3.65_{\pm 0.3}$ & $3.22_{\pm 0.4}$ & $3.39_{\pm 0.5}$ & $4.05_{\pm 0.5}$ \\
\rowcolor{lightred} GALA   & $17.27_{\pm 0.1}$ & $17.79_{\pm 0.7}$ & $22.21_{\pm 0.1}$ & $22.37_{\pm 0.1}$ \\
\hline
\end{tabular}
}
\label{tab:runtime}
\end{table}

For KD3A, while initial source training is parallelized, the subsequent consensus and knowledge-voting stages are executed sequentially on the target \citep{kd3a}, forcing all sources to remain idle during this phase. In contrast, GALA and FACT parallelize all local training and fine-tuning steps. We report the maximum per-source runtime per round as a practical upper bound, averaged over five runs to mitigate hardware fluctuations. As shown in the results, KD3A scales poorly because its consensus step requires computing source permutations, leading to exponential runtime growth as the number of domains increases. GALA and FACT avoid this bottleneck, maintaining high efficiency. While GALA incurs a slightly higher per-round cost than FACT due to the inter-group alignment, it avoids the computational infeasibility seen in  KD3A in diverse, large-scale multi-domain scenarios.

\subsection{Parameter Analysis: $\tau$ in MDMGB+}



\begin{figure}[ht]
    \centering
    \includegraphics[width=\linewidth]{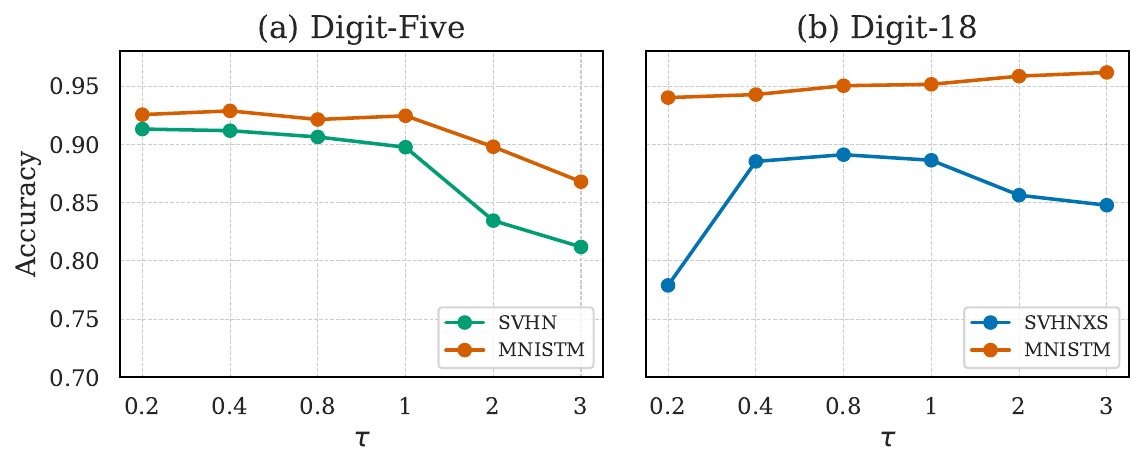}
    \caption{Effect of $\tau$ in GALA on performance for SVHN and MNIST-M (Digit-Five) and SVHNXS and MNIST-M (Digit-18).}
    \label{fig:parameter_analysis}
\end{figure}

We evaluate the sensitivity of MDMGB+ to the temperature parameter $\tau$, which controls the sharpness of source relevance weighting. Figure~\ref{fig:parameter_analysis} reports the final target accuracy (mean over five runs) for different values of $\tau$, with (a) corresponding to Digit-Five and (b) to Digit-18 sources.

On Digit-18, very low temperatures (e.g., $\tau=0.2$) produce overly uniform weights, limiting the model’s ability to focus on well-aligned sources, as reflected by reduced accuracy. In contrast, excessively high values (e.g., $\tau=3$) result in lower final performance despite initially faster convergence. Intermediate values ($\tau \in [0.8, 1.0]$) offer the best trade-off, yielding stable and accurate performance.

On Digit-Five, which exhibits lower source diversity, lower to moderate temperatures ($\tau \in [0.4, 1.0]$) achieve the highest accuracy, while larger values tend to degrade performance. This behavior is consistent with the fact that many source domains are relevant in this setting, and overly sharp weighting can unnecessarily overemphasize a small subset.

Overall, these results suggest that moderate $\tau$ values provide an effective default across benchmarks, while additional knowledge about source diversity or access to validation data can be leveraged to further fine-tune $\tau$ for additional performance gains.

\subsection{Ablation Study}



To isolate the contribution of our method, we evaluate the following configurations of GALA under the full 17-source Digit-18 setup: (1) IGD with uniform source contributions (no weighting), (2) IGD combined with MDMGB \citep{SFDA}, (3) IGD combined with our proposed MDMGB+ with $\tau=1.0$, (4) GALA without target training (no IGD) combined with MDMGB+ ($\tau=1.0$), and (5) GALA without target training and without weighting.

\begin{figure}[ht]
    \centering
    \includegraphics[width=\linewidth]{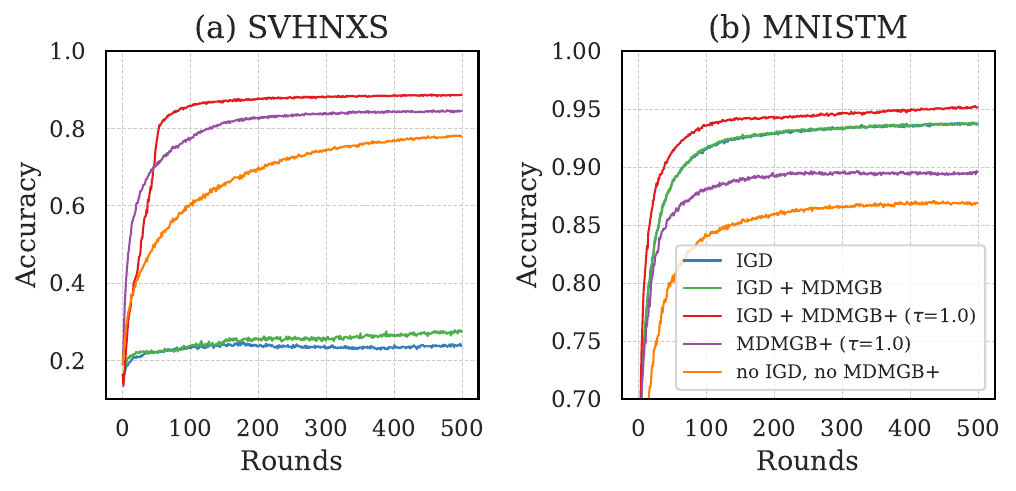}
    \caption{Effect of IGD and weighting strategies (no weighting, MDMGB, MDMGB+) (Digit-18).}
    \label{fig:ablation_study}
\end{figure}

Figure~\ref{fig:ablation_study} reports results for two challenging targets, MNISTM and SVHNXS. On MNISTM, each component of GALA contributes positively, with IGD being more important than weighting. On SVHNXS, by contrast, domain selection with MDMGB+ is crucial. IGD alone fails to converge, MDMGB offers almost no improvement, and variants without MDMGB+ remain weaker. Overall, the full combination of IGD and MDMGB+ yields the most stable convergence and strongest performance by effectively capturing domain relevance under high source diversity.

\subsection{Limitations}
\label{sec:limitations}

While GALA scales effectively to many source domains, it incurs higher computational and communication costs per round since all sources participate in training and fine-tuning. This full participation, though parallelizable, contrasts with more selective methods like FACT. Future work could reduce source participation or communication frequency to improve efficiency. Although our experiments focus on digit datasets, we introduce Digit-18 to address the scarcity of public benchmarks with many diverse sources. Extending evaluation to broader domains remains an important direction.

\section{Conclusion} \label{conclusion}

We introduced GALA, a federated framework for unsupervised multi-source domain adaptation that addresses the scalability challenges of diverse multi-source settings. By combining temperature-scaled centroid-based weighting with inter-group discrepancy minimization, GALA enables robust, efficient alignment of diverse source domains to an unlabeled target. Our method achieves state-of-the-art performance across standard UMDA benchmarks and demonstrates strong stability and accuracy in large-scale settings where existing approaches degrade or fail to converge. Through our new Digit-18 benchmark, we further validate GALA's effectiveness under realistic, high-diversity conditions.

\section*{Impact Statement}

This paper presents work whose goal is to advance the field of Machine
Learning. There are many potential societal consequences of our work, none
which we feel must be specifically highlighted here.

\bibliography{references}
\bibliographystyle{icml2026}

\newpage
\appendix
\onecolumn

\section*{Appendix}
\section{Intuition: Connection between Divergence in Corollary 1 and the $\mathcal{L}$s}\label{app:connection}
Here we make explicit how the divergence term in Proposition \ref{thm:multisource} relates to \(\mathcal{L}_{\mathrm{full}}, \mathcal{L}_{\mathrm{IDD}}^{(i,j)}\) and \(\mathcal{L}_{\mathrm{IGD}}\). Recalling the definition of $d_{\mathcal{H}\Delta\mathcal{H}}(\mathbb{D}_S,\mathbb{D}_T)$ for binary classification:

\begin{equation}
    2 \sup_{h,h' \in \mathcal{H}} \left| \Pr_{x \sim \mathbb{D}_S}[h(x) \neq h'(x)] - \Pr_{x \sim \mathbb{D}_T}[h(x) \neq h'(x)] \right|.
\end{equation}

Intuitively, over pairs $h,h'\in\mathcal H$ we (i) find the region where they disagree, (ii) measure the mass that $D_S$ and $D_T$ place on that region, and (iii) take the largest (absolute) difference.

To obtain a practical, empirical estimator we restrict $\mathcal H$ to the classifiers we actually have at a given round (i.e $F_1\circ G,\dots,F_N\circ G$). For multiclass tasks we relax the binary disagreement to the $ell_1$ difference of predicted probability vectors. Under this restriction, we can see that IDD (Eq. \ref{eq:idd}) captures the same target-side disagreement notion as $\Pr_{x\sim\mathbb{D}_T}[h(x)\neq h'(x)]$.

Prior DA work \citep{saito2018maximum} often assumes that source-side disagreement is negligible because each \(F_i\circ G\) is well-trained on its own source, i.e. $\Pr_{x\sim\mathbb{D}_S}[h(x)\neq h'(x)]\approx 0$. We do not require this strong assumption because of source heterogeneity and instead we make the milder empirical assumption that a well-trained $G$ allows $\Pr_{x \sim \mathbb{D}_S}[h(x) \neq h'(x)] \ll \Pr_{x \sim \mathbb{D}_T}[h(x) \neq h'(x)]$, which we find holds empirically. Under this assumption the divergence \(d_{\mathcal{H}\Delta\mathcal{H}}\) is dominated by target-side disagreement, so minimizing target-side disagreement terms provides an effective proxy for the divergence term in Corollary \ref{thm:multisource}.

\section{Intuition: IGD as a biased but low-variance estimator of $\mathcal{L}_{\mathrm{full}}$}\label{app:intuition}
IGD and IDD both aim to estimate $\mathcal{L}_{\mathrm{full}}$, which is expensive to compute. IDD is an unbiased but high-variance estimator of $\mathcal{L}_{\mathrm{full}}$. We compare the bias and variance of IGD to those of IDD.

\subsection{IGD has Lower Variance than IDD}

Using the group predictors defined in Eq.~\ref{eq:group_predictors}, with $\sum_{i\in G_k}\tilde w_i = 1$ and $\tilde w_i \ge 0$, the IGD objective can be written as

\begin{align*}
    \mathcal{L}_{\mathrm{IGD}}&= \mathbb{E}_{x \sim \mathbb{D}_T}\left[ \| F_{\mathcal{G}_1}(G(x)) - F_{\mathcal{G}_2}(G(x)) \|_1 \right]\\
    &= \mathbb{E}_{x \sim \mathbb{D}_T}\left[ \left\|\sum_{i\in\mathcal{G}_1} \tilde{w}_iF_i(G(x)) - \sum_{j\in\mathcal{G}_2} \tilde{w}_jF_j(G(x)) \right \|_1 \right]\\
    &=\mathbb E_{x\sim\mathcal D_T}\left[\left\|\sum_{i\in G_1}\sum_{j\in G_2}\tilde w_i\tilde w_j (F_i(G(x)) -F_j(G(x)))\right\|_1\right].
\end{align*}

Because IGD averages many pairwise terms before taking the norm, its variance across random splits is smaller than that of an estimator that uses only a single pair $(i,j)$ per update, provided the weights are non-adversarial. Furthermore, Jensen's inequality yields

\begin{equation*}
   \mathcal{L}_{\mathrm{IGD}}\leq \sum_{i\in G_1}\sum_{j\in G_2}
\tilde w_i\tilde w_j E_{x\sim\mathcal D_T}\left[\left\|F_i(G(x))-F_j(G(x))\right\|_1\right] = \sum_{i\in G_1}\sum_{j\in G_2} \tilde w_i\tilde w_j\ \mathcal{L}_{\mathrm{IDD}}^{(i,j)} \le \max_{i,j}\mathcal{L}_{\mathrm{IDD}}^{(i,j)}.
\end{equation*}

While the observations above hold in general, we make the variance reduction explicit under a simpler setting. \underline{For exposition}, assume we have even $N=2M$ source predictors $F_1,\dots,F_{2M}$ and form a random balanced partition into two groups $G_1,G_2$ of size $M$ each. Further assume uniform group-internal weights $\tilde w_i=1/M$ for $i\in G_k$. Finally, let's assume binary classification.

We will decompose the $\ell_1$ norm as sum of coordinates. We begin by noticing that there will be one coordinate of $F_i(G(x))-F_j(G(x))$ that is in $[0, 1]$. We denote that variable $h_a(x) -h_b(x)$, where $a, b \in \{i, j\}$. Naturally the remaining coordinate would be $(1-h_a(x)) - (1-h_b(x)) =h_b(x) -h_a(x) \in [-1, 0]$, because $F$ maps to a probability vector. Further denote $\bar{h}_{G}(x) =\sum_{i\in G}h_i(x)$, and $\bar{h}(x)$ the average of all $h_i(x)$. 

We aim to compare the following quantities:
\begin{align}
    \text{Variance of IDD:} ~&~ \mathbb{V}_{a,b} \left[\mathbb{E}_x\left [ 2(h_{a}(x)-h_b(x)) \right ]\right] \approx 8\mathbb{V}_{k} \left[\mathbb{E}_x\left [ h_{k}(x)\right ]\right], \label{eq:V_IDD} \\
    \text{Variance of IGD:} ~&~ \mathbb{V}_{G_1}\left[\mathbb{E}_x \left[ 2(\bar{h}_{G_1}(x) - \bar{h}_{G_2}(x))\right]\right]=\mathbb{V}_{G_1}\left[\mathbb{E}_x \left[ 2(2\bar{h}_{G_1}(x) - 2\bar{h}(x)) \right]\right].
\end{align}
Then, 
\begin{align}
     \text{Variance of IGD} &=  16\mathbb{V}_{G_1}\left[ \mathbb{E}_x \left[ \bar{h}_{G_1}(x)  \right]\right]&\text{since } \mathbb{E}_x\bar{h}(x) \text{ is a constant w.r.t. } G_1\\
    &=16\mathbb{V}_{G_1}\left[\frac{1}{M}\sum_{i\in G_1} \mathbb{E}_x \left[ h_{i}(x)  \right]\right] &\text{by linearity of expectation}\\
     &=16\cdot \frac{2M-M}{2M-1}\frac{\mathbb{V}_{k}\left[\mathbb{E}_x \left[ h_{k}(x)  \right]\right]}{M}&\text{variance of sample mean without replacement}\\
     &\approx \frac{8}{M}\mathbb{V}_{k}\left[\mathbb{E}_x \left[ h_{k}(x)  \right]\right]\\
     &\approx \frac{1}{M}\text{Variance of IDD}.
\end{align}

Experimentally, we verify this $1/M$ scaling behavior under two toy settings: 1) $h_{i}(x) \sim Unif[0,1]$, and 2) $\mu_{i} \sim Unif[\theta,1-\theta]$, then $h_i(x) \sim Unif[\mu_i-\theta, \mu_i + \theta]$.

\subsection{IGD is Biased under Uniform Weights}
Again \underline{for exposition}, assume we have even $N=2M$ source predictors $F_1,\dots,F_{2M}$ and form a random balanced partition into two groups $G_1,G_2$ of size $M$ each. Further assume uniform group-internal weights $\tilde w_i=1/M$ for $i\in G_k$. Define the normalized full pairwise discrepancy:
\begin{equation*}
\text{(normalized)}\mathcal{L}_{\mathrm{full}}= \frac{1}{M(2M-1)}\sum_{i<j} \mathbb E_{x\sim\mathcal D_T}\left[\|F_i(G(x)) - F_j(G(x))\|_1\right].
\end{equation*}
Taking expectation over the random split gives:

\begin{equation*}
    \mathbb E_{\text{split}}[\mathcal{L}_{\mathrm{IGD}}]\le \mathbb E_{\text{split}}\left[\frac{1}{M^2} \sum_{i\in G_1}\sum_{j\in G_2} \mathcal{L}_{\mathrm{IDD}}^{(i,j)}\right].
\end{equation*}

For any unordered pair $(i,j)$ the probability they fall into different groups is $P_{ij}=\frac{M}{2(2M-1)}$. Therefore

\begin{equation*}
    \begin{split}
        \mathbb E_{\text{split}}[\mathcal{L}_{\mathrm{IGD}}]\le \mathbb E_{\text{split}}\left[\frac{1}{M^2} \sum_{i\in G_1}\sum_{j\in G_2}\mathcal{L}_{\mathrm{IDD}}^{(i,j)}\right] = \frac{1}{M^2}\sum_{i\ne j}P_{ij} \cdot \mathcal{L}_{\mathrm{IDD}}^{(i,j)} \\
        = \frac{1}{M^2} \cdot 2\sum_{i<j}\frac{N}{2(2M-1)}\mathcal{L}_{\mathrm{IDD}}^{(i,j)} = \text{(normalized) }\mathcal{L}_{\mathrm{full}},
    \end{split}
\end{equation*}

so IGD (in expectation under uniform weights and balanced splits) underestimates the normalized full-pairwise objective. GALA's strong empirical performance suggests that introducing this bias was worth the gain in reduced variance compared to $L_{IDD}^{(i,j)}$.

\section{GALA: Algorithm Overview}
\label{sec:GALA-overview}

We next describe the proposed GALA framework. The complete procedure is detailed in Algorithm~\ref{alg:GALA}.Each round begins with the server broadcasting the global model $(G_t, F_t)$ to all domains. Domains compute class-wise centroids and upload them to the server, which calculates normalized relevance scores via MDMGB+ to weight each source's contribution. Sources update their models locally using cross-entropy loss. The server aggregates feature extractors with similarity-based weights to form a shared extractor $G'$, sent back to sources. With $G'$ frozen, sources fine-tune predictors and return updates to the server. The server randomly partitions predictors into two groups, averages them, and sends both groups with $G'$ to the target. The target updates $G'$ to $G''$ by minimizing IGD loss between group predictions. The server merges the group predictors into a global predictor $F_{t+1}$ and sets $G_{t+1} \leftarrow G''$, updating the model as $h_{t+1} = F_{t+1} \circ G_{t+1}$. This completes one round.

\begin{algorithm}[h]
   \caption{Training Process of GALA}
   \label{alg:GALA}
\begin{algorithmic}
   \STATE {\bfseries Input:} Source datasets $\{\mathbb{D}_S^n\}_{n=1}^N$, target dataset $\mathbb{D}_T$, initial model $(G, F)$, total rounds $T$, temperature $\tau$
   
   \FOR{$t = 1$ {\bfseries to} $T$}
      \STATE Broadcast global model $(G_t, F_t)$ to all domains
      
      \FOR{each class $c \in C$}
         \STATE {\bfseries in parallel for each source $n$:} 
         \STATE Compute $r_n^c \leftarrow \frac{\sum_{x \in \mathbb{D}_S^n} \delta_c(x) G(x)}{\sum_{x \in \mathbb{D}_S^n} \delta_c(x)}$
         \STATE Target computes $r_T^c \leftarrow \frac{\sum_{x \in \mathbb{D}_T} \delta_c(x) G(x)}{\sum_{x \in \mathbb{D}_T} \delta_c(x)}$
      \ENDFOR
      
      \STATE Compute domain similarity $S(r_T, r_n)$ via Eq.~\ref{eq:domain_similarity}
      \STATE $w_n \leftarrow \texttt{MDMGBPlus}(r_T, r_n)$ via Eq.~\ref{eq:mdmgb+}
      
      \FOR{each source $n$ {\bfseries in parallel}}
         \STATE Initialize $(G_n, F_n) \gets (G_t, F_t)$
         \STATE Update $(G_n, F_n)$ by optimizing $\mathbb{E}_{(x,y)\sim \mathbb{D}_S^n}[\ell(F_n(G_n(x)),y)]$
      \ENDFOR
      
      \STATE Aggregate $G' \leftarrow \sum_n w_n G_n$ and broadcast to sources
      
      \FOR{each source $n$ {\bfseries in parallel}}
         \STATE Freeze $G'$, fine-tune $F_n$ on $\mathbb{D}_S^n$
         \STATE Send updated $F_n$ to server
      \ENDFOR
      
      \STATE Randomly split sources into groups $\mathcal{G}_1$ and $\mathcal{G}_2$
      
      \FOR{each group $\mathcal{G}_i \in \{\mathcal{G}_1, \mathcal{G}_2\}$}
         \FOR{each source $n \in \mathcal{G}_i$}
            \STATE Compute normalized weight $\tilde{w}_n$ via Eq.~\ref{eq:normalized_weights}
         \ENDFOR
         \STATE $F_{\mathcal{G}_i} \leftarrow \sum_{n \in \mathcal{G}_i} \tilde{w}_n F_n$
         \STATE $w_{\mathcal{G}_i} \leftarrow \sum_{n \in \mathcal{G}_i} w_n$
      \ENDFOR
      
      \STATE Send $(F_{\mathcal{G}_1}, F_{\mathcal{G}_2}, G')$ to target domain
      \STATE Target updates $G'$ to $G''$ by minimizing $\mathcal{L}_{\mathrm{IGD}}$
      \STATE $G_{t+1} \gets G''$
      \STATE Aggregate global predictor $F_{t+1} \leftarrow w_{\mathcal{G}_1} F_{\mathcal{G}_1} + w_{\mathcal{G}_2} F_{\mathcal{G}_2}$
      
   \ENDFOR
\end{algorithmic}
\end{algorithm}

\section{Implementation Details} \label{sec:implementation}

We provide here the complete architectural specifications and training hyperparameters for reproducibility. The source code to reproduce all experiments will be made publicly available upon publication.

\subsection{Architectures}

For experiments on Digit-Five and Digit-18, we use a lightweight 2-layer CNN as the feature extractor, followed by a 3-layer MLP predictor. The full architecture is detailed in Table~\ref{tab:cnn_arch}. For Office-Caltech10, we adopt a ResNet101 backbone pretrained on ImageNet, followed by a task-specific MLP predictor as outlined in Table~\ref{tab:resnet_head}.

\begin{table*}[ht]
\centering
\caption{Digit datasets Model Architecture} \label{tab:cnn_arch}
\begin{tabular}{cccc}
\hline
\textbf{Layer} & \textbf{Output Size} & \textbf{Kernel / Units} & \textbf{Details} \\
\hline
Input & $3 \times 32 \times 32$ & - & RGB Image \\
Conv2D + BN + ReLU & $64 \times 32 \times 32$ & $5 \times 5$ & padding=2 \\
MaxPool2D & $64 \times 16 \times 16$ & $3 \times 3$ & stride=2, padding=1 \\
Conv2D + BN + ReLU & $128 \times 16 \times 16$ & $5 \times 5$ & padding=2 \\
MaxPool2D & $128 \times 8 \times 8$ & $3 \times 3$ & stride=2, padding=1 \\
Flatten & $8192$ & - & \\
\hline
Dropout + FC + BN + ReLU & $3072$ & - & p=0.5 \\
Dropout + FC + BN + ReLU & $100$ & - & p=0.5 \\
Dropout + FC + BN + Softmax & $10$ & - & p=0.5 \\
\hline
\end{tabular}
\end{table*}

\begin{table*}[ht]
\centering
\caption{ResNet-based Predictor Architecture} \label{tab:resnet_head}
\begin{tabular}{cccc}
\hline
\textbf{Layer} & \textbf{Output Size} & \textbf{Units} & \textbf{Details} \\
\hline
ResNet101 Backbone & $1000$ & - & Pretrained on ImageNet \\
\hline
Dropout + FC + BN + ReLU & $500$ & - & p=0.5 \\
FC + BN + Softmax & $\{10, \text{Number of Classes}\}$ & - & Task-specific classes \\
\hline
\end{tabular}
\end{table*}

\subsection{Training Details}

Table~\ref{tab:impl_params} summarizes the training parameters used for each benchmark. All models are trained using SGD with momentum 0.9 and weight decay $5 \times 10^{-4}$. We set the batch size to 128 and train for 500 rounds. Communication occurs once per round ($r = 1$), and each training phase (source training, fine-tuning, adversarial alignment) is performed for one epoch. Following \cite{kd3a}, we apply mixup augmentation ($\alpha = 0.2$) for Office-Caltech10 only.

\begin{table*}[ht!]
\centering
\caption{Implementation details of our GALA on three benchmark datasets.} \label{tab:impl_params}
\begin{tabular}{c|c|c|c}
\hline
\textbf{Parameters} & \textbf{Digit-Five} & \textbf{Digit-18} & \textbf{Office-Caltech10} \\
\hline
Data Augmentation &\multicolumn{2}{c|}{None} & Mixup ($\alpha = 0.2$) \\
\hline
Backbone &\multicolumn{2}{c|}{2-layer CNN} & ResNet101 (pretrained=True) \\
\hline
Optimizer & \multicolumn{3}{c}{SGD with momentum = 0.9 and weight decay =$5 \times10^{-4}$} \\
\hline
Learning Rate Schedule & \multicolumn{2}{c|}{CustomLR ($\gamma$=0.75)} & ExponentialLR ($\gamma$=0.9) \\
\hline
Batch Size &\multicolumn{3}{c}{128} \\
\hline
Total Rounds & \multicolumn{3}{c}{500} \\
\hline
Communication Rounds & \multicolumn{3}{c}{$r = 1$} \\
\hline
Temperature & $\tau=0.2$ & \multicolumn{2}{c}{$\tau=1.0$} \\
\hline
\end{tabular}
\end{table*}

\subsection{Hardware}

All experiments were run on a compute node with an AMD EPYC 7713 64-core CPU and a single NVIDIA A100 GPU (40GB).

\section{Datasets}

\paragraph{Office-Caltech10. } Office-Caltech10 consists of for domains:  Amazon, Webcam, DSLR and Caltech. The images show objects from 10 different classes which are shared between Office \cite{OfficeDataset} and Caltech-265 \cite{Caltech256} datasets. 

\paragraph{Digit-Five. } The Digit-Five dataset \cite{zhao2020multi} is a popular benchmark for digit recognition. It consists of the following five datasets, each representing a separate domain: MNIST, MNIST-M, Street-View House Numbers (SVHN), Synthetic Digits (SYN), and USPS.

\subsection{Digit-18 Benchmark} \label{sec:digit-18}

\textbf{Digit-18} is our proposed large-scale benchmark composed of 18 domains, created by applying systematic transformations to existing digit datasets. It is specifically designed to evaluate the robustness and scalability of UMDA methods in high-source scenarios. Our goal was to ensure sufficient variability and domain shifts across the domains. Thus, we did not apply each transformation to every dataset, as some domains are already similar. Full dataset will be made available upon publication.

\begin{figure*}[ht!]
\centering
\includegraphics[width=\textwidth]{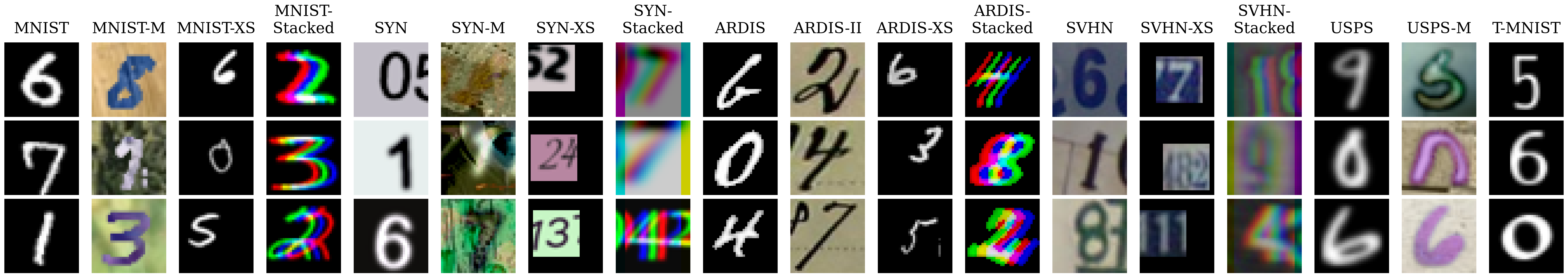}
\caption{Sample images from each domain in the \textbf{Digit-18} benchmark.}
\label{fig:digit18_samples}
\end{figure*}

\subsubsection{Base Datasets}

\begin{itemize}
    \item \textbf{ARDIS}~\cite{ARDIS}: A historical handwritten digit dataset extracted from Swedish church records. We use 6,600 training and 1,000 testing samples. We include two variants: a normalized version (matched to MNIST) and an unprocessed version with original grayscale backgrounds and image noise, referred to as \textbf{ARDIS II}.
    
    \item \textbf{TMNIST}~\cite{TMNIST}: Typography-MNIST contains 22,400 training and 7,500 test images of digits rendered in various fonts. The images are grayscale on a black background, similar to MNIST but with greater stylistic diversity.
\end{itemize}

\subsubsection{Domain Transformations}

We applied the following transformation strategies to simulate diverse and challenging domain shifts:

\begin{itemize}
    \item \textbf{Background Augmentation:} Following MNIST-M~\citep{MNISTM}, we overlay complex colored backgrounds on digit images from SYN, SVHN, and USPS to create SYNM, SVHNM, and USPSM.
    
    \item \textbf{Scaling:} Original digit images are resized to $20\times20$ and re-centered on a $32\times32$ black canvas. Applied to MNIST, SYN, SVHN, and ARDIS, yielding *-XS domains (e.g., MNISTXS).
    
    \item \textbf{Stacking:} We introduce pixel-level channel misalignments by shifting R, G, B channels in opposite directions. Applied to grayscale domains, this generates color interference effects. Used for MNIST, SYN, SVHN, and ARDIS to generate *-STACK domains.
\end{itemize}

\subsubsection{Domain Shift Analysis}

\begin{figure*}[ht!]
    \centering
    \includegraphics[width=0.9\textwidth]{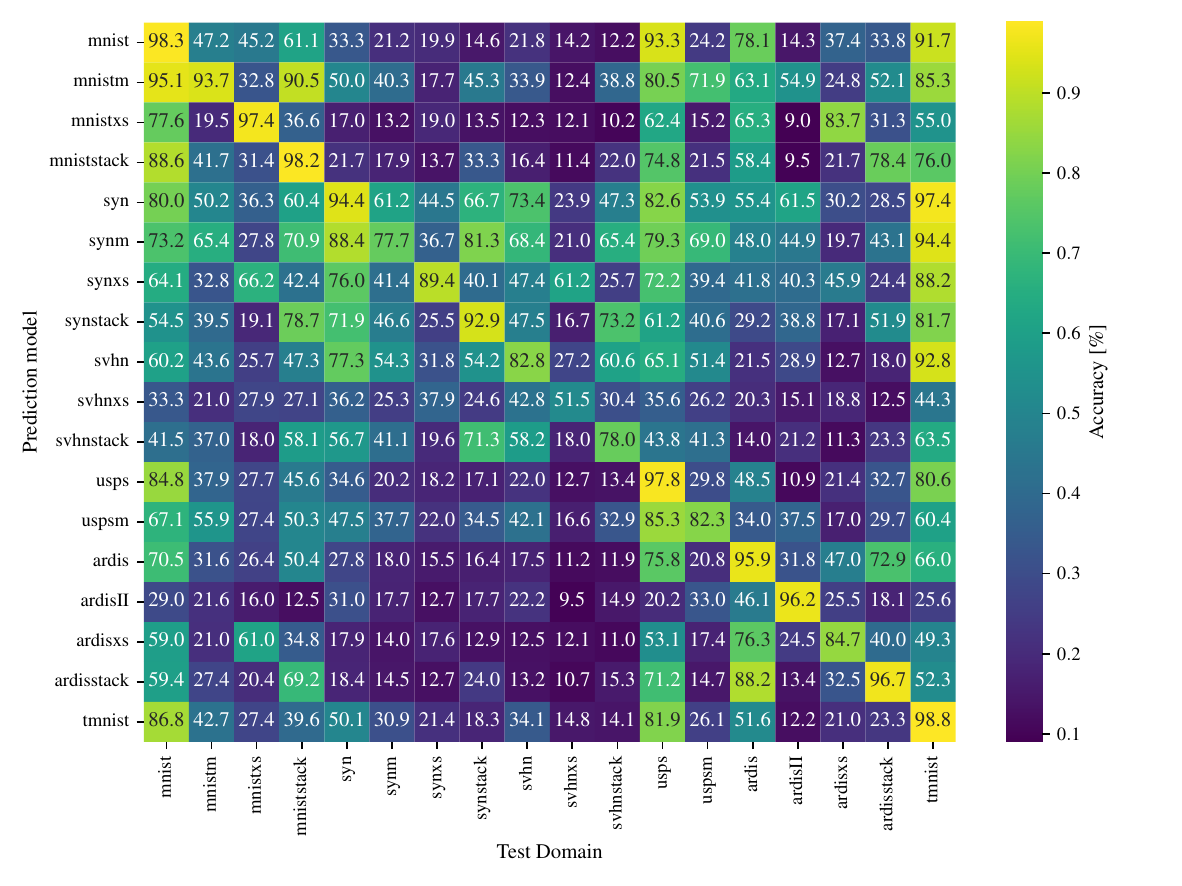}
    \caption{Cross-domain similarity matrix: each row corresponds to a model trained on a source domain and evaluated on all target domains.}
    \label{fig:similarity_heatmap}
\end{figure*}

To assess domain similarity and difficulty, we trained simple models on each domain independently and evaluated them across all other domains. These models used the same architecture and training settings as in the UFDA experiments (500 epochs, SGD with momentum 0.9, fixed learning rate 0.001). The accuracy matrix in Figure~\ref{fig:similarity_heatmap} reveals cross-domain generalization trends and helps characterize inter-domain shifts.

Notably, models trained on clean datasets (e.g., MNIST) fail to generalize well to complex variants (e.g., SYNM), while models trained on background-augmented domains (e.g., MNISTM) transfer better to simpler settings. These insights informed the domain selection process and help contextualize results in our experiments.

\section{Communication-cost analysis}
\label{app:comm-cost}

In this work, we focus on the algorithmic scalability and stability of the distributed UMDA objective under many sources. Practical FL networking issues (client dropouts, bandwidth limits, etc.) are independent challenges, and we follow the same standard assumptions used in most FL literature, where all clients participate in each round.

Importantly, GALA does not introduce heavy additional communication. Each source sends only a centroid vector for each class $C$, which is negligible in size. We also performed an additional communication-cost analysis. The total upload/download cost per round (approx.) is reported in Table~\ref{tab:comm-cost}.

\begin{table*}[ht]
\centering
\caption{Total upload/download cost per round (approx.).}
\begin{tabular}{c|ccc|ccc}
\hline
\rowcolor{lightgray} Method & \multicolumn{3}{c|}{5 Sources} & \multicolumn{3}{c}{10 Sources} \\
\rowcolor{lightgray} & Source & Server & Target & Source & Server & Target \\
\hline
FACT  & $\sim 205\ \mathrm{MB}$ & $\sim 618\ \mathrm{MB}$ & $\sim 205\ \mathrm{MB}$   & $\sim 205\ \mathrm{MB}$ & $\sim 618\ \mathrm{MB}$  & $\sim 205.63\ \mathrm{MB}$ \\
GALA  & $\sim 207\ \mathrm{MB}$ & $\sim 1.24\ \mathrm{GB}$ & $\sim 205\ \mathrm{MB}$   & $\sim 207\ \mathrm{MB}$ & $\sim 2.27\ \mathrm{GB}$  & $\sim 205.63\ \mathrm{MB}$ \\
KD3A  & $\sim 205\ \mathrm{MB}$ & N/A                     & $\sim 1.03\ \mathrm{GB}$ & $\sim 205\ \mathrm{MB}$ & N/A                      & $\sim 2.05\ \mathrm{GB}$ \\
\hline
\end{tabular}
\label{tab:comm-cost}
\end{table*}

Overall, FACT is the most communication-efficient due to its selective strategy. Although GALA involves all sources every round, its total communication cost remains comparable to KD3A while providing a centralized FL workflow and not requiring the target to receive all source models.

\section{Robustness Analysis of FACT}
\label{appendix:fact-robustness}

\begin{figure*}[ht]
    \centering
    \includegraphics[width=\textwidth]{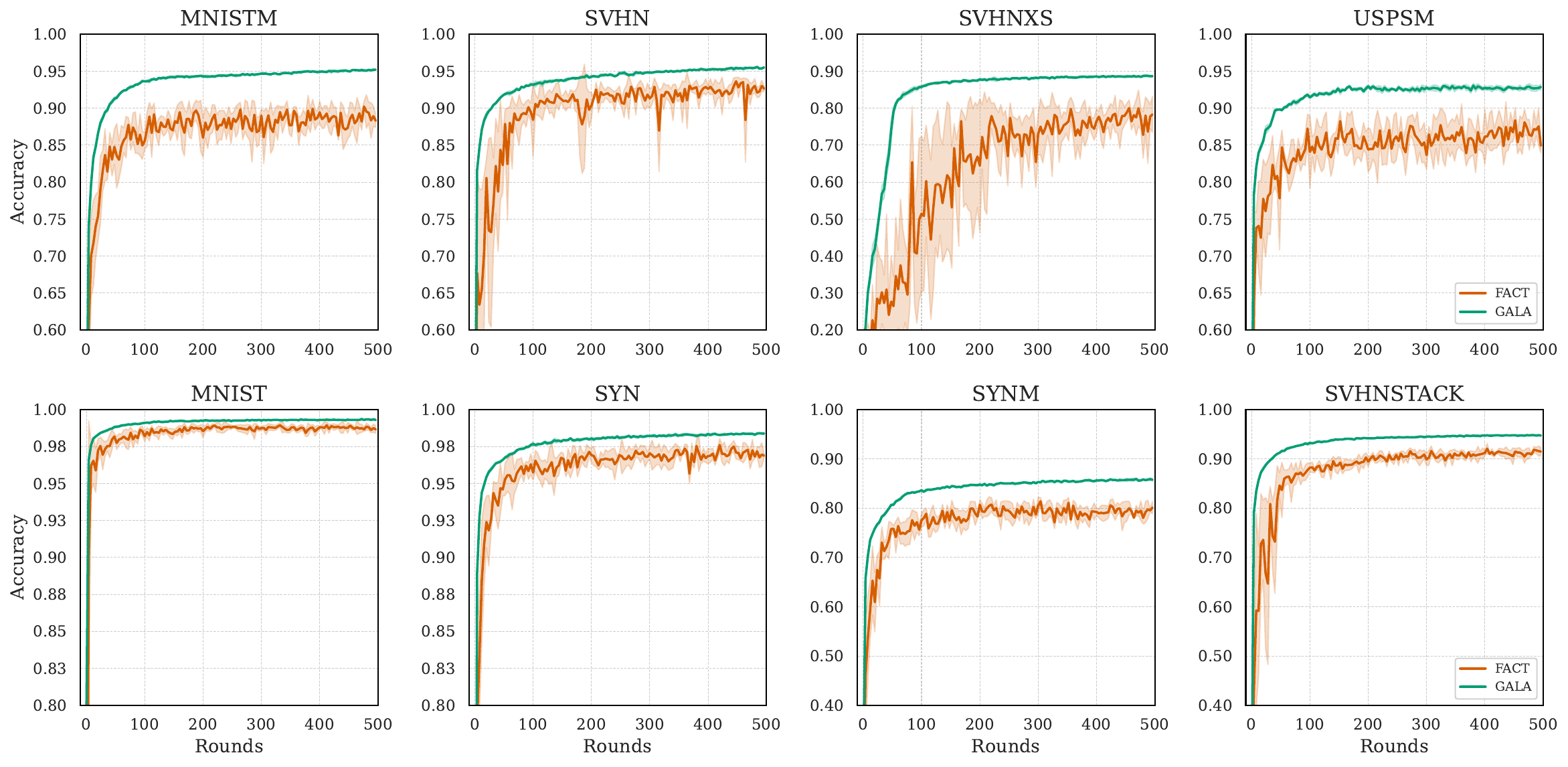}
    \caption{Test accuracy over training rounds for different target domains in the full Digit-18 setup (extension of Figure~\ref{fig:digit18_performance_1}).}
    \label{fig:extended_convergence_plot}
\end{figure*}

\begin{figure*}[ht]
    \centering
    \includegraphics[width=0.6\textwidth]{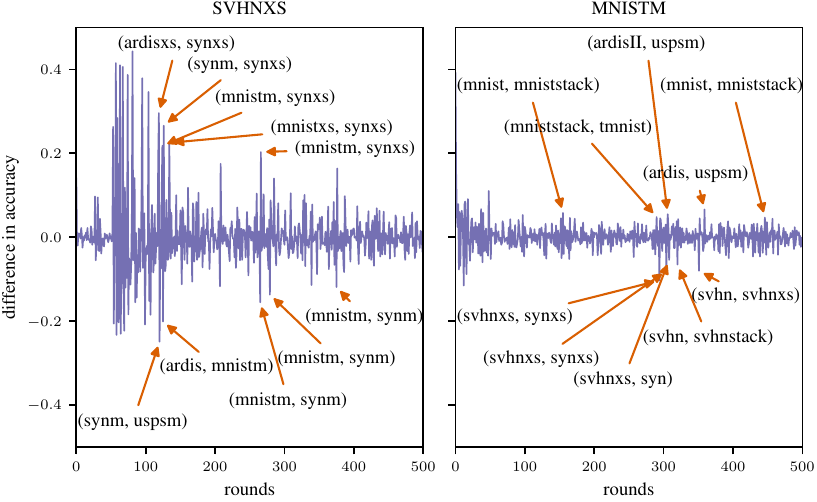}
    \caption{Round-to-round accuracy difference of FACT. Clients associated with the highest single-round accuracy increases and decreases are annotated.}
    \label{fig:FACT_drops}
\end{figure*}

FACT randomly selects two source domains in each communication round to perform inter-domain distance minimization. This random pairing strategy introduces instability during training, as model updates become highly sensitive to the particular source combination selected in each round. As a result, we observe frequent fluctuations in test accuracy, especially on more challenging target domains, as illustrated in Figure~\ref{fig:extended_convergence_plot}.
To examine this behavior in more detail, we analyze training dynamics on the Digit-18 benchmark. Figure~\ref{fig:FACT_drops} shows round-to-round changes in test accuracy for two particularly challenging target domains, SVHNXS and MNISTM. For interpretability, we annotate the source domain pairs selected in rounds exhibiting the largest single-round accuracy increases and decreases. To focus on convergence behavior, we restrict this analysis to the phase after the first 110 communication rounds (i.e., after the warm-up period).

\begin{figure*}[ht]
    \centering
    \includegraphics[width=0.7\textwidth]{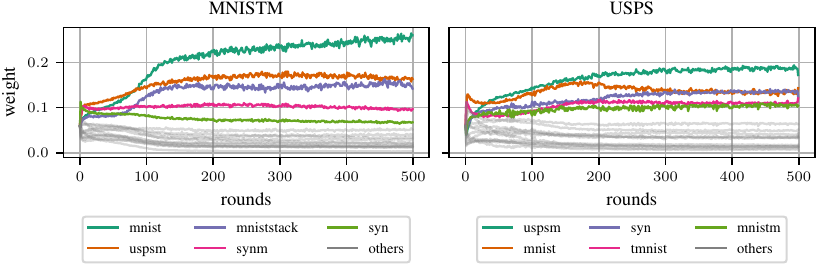}
    \caption{Evolution of source weights assigned by MDMGB+ for MNISTM and USPS (Digit-18 setting). The five most highly weighted source domains are highlighted. Shifts over time reflect both domain similarity and the model’s adaptation to target-specific learning needs.}
    \label{fig:dynamic_weights}
\end{figure*}

For SVHNXS, the largest accuracy drops occur when both selected sources are *-M domains, all of which exhibit similarity scores below 25\% with SVHNXS (see similarity matrix in Figure~\ref{fig:similarity_heatmap}). Since SVHNXS consists of black-and-white digit images with extensive black backgrounds, it cannot effectively leverage the colorful backgrounds characteristic of *-M sources, leading to negative transfer. In contrast, pairings such as MNISTM with SYNXS result in marked performance improvements, indicating that a dissimilar source can still be beneficial when combined with a complementary one. Several *-XS sources, particularly SYNXS, consistently produce significant accuracy gains. This behavior can be attributed to similar data generation processes across these domains, which enable the model to better capture the characteristics of SVHNXS. Notably, SYNXS, the most similar source to SVHNXS, appears in all beneficial source pairs. We further observe that SYNXS is consistently assigned the highest weight by GALA, demonstrating our method’s ability to identify and emphasize the most relevant sources.

A similar pattern emerges for MNISTM: selecting MNIST-like and -M domains typically leads to improvements in test accuracy relative to the previous round, whereas selecting SVHN- domains and SYNXS, which have similarity scores below 50\%, often results in substantial accuracy drops.

Overall, these results highlight the robustness limitations of FACT’s random source selection mechanism in large-scale multi-source settings.

\section{Adaptive Source Weighting in GALA}
\label{appendix:GALA-weights}

GALA dynamically assigns a weight to each source client in every training round, determining its influence on the shared model. Figure \ref{fig:dynamic_weights} illustrates the evolution of these weights for the target domains MNISTM and USPS in the Digit-18 setting. The top five most highly weighted sources are highlighted.

For MNISTM, the top-weighted domains align with those found most similar in our similarity analysis. These include MNIST-like and *-M datasets, supporting the expectation that their combination is well-suited for learning MNISTM. The USPS plot illustrates the value of dynamic re-weighting. In early rounds, simpler domains like MNIST and TMNIST dominate. Later, the weights shift toward USPSM and SYN, reflecting a focus on learning finer details and USPS-specific features.

\end{document}